\documentclass[runningheads]{llncs}
\usepackage[margin=1in]{geometry}
\usepackage[T1]{fontenc}
\usepackage{graphicx}
\usepackage{amsmath} 
\usepackage{booktabs}
\usepackage{marvosym}
\usepackage{amsfonts}
\usepackage{algorithm} 
\usepackage{algpseudocode} 

\begin{document}
\title{E(3)-invariant diffusion model for pocket-aware peptide generation}
%
%
\author{Po-Yu Liang \and Jun Bai\textsuperscript{\Letter}}
%
%
\institute{
    Department of Computer Science, University of Cincinnati, Ohio, United State\\
    \email{liangpu@mail.uc.edu, baiju@ucmail.uc.edul}
}
\maketitle              
\begin{abstract}
Biologists frequently desire protein inhibitors for a variety of reasons, including use as research tools for understanding biological processes and application to societal problems in agriculture, healthcare, etc. Immunotherapy, for instance, relies on immune checkpoint inhibitors to block checkpoint proteins, preventing their binding with partner proteins and boosting immune cell function against abnormal cells. Inhibitor discovery has long been a tedious process, which in recent years has been accelerated by computational approaches. Advances in artificial intelligence now provide an opportunity to make inhibitor discovery smarter than ever before. While extensive research has been conducted on computer-aided inhibitor discovery, it has mainly focused on either sequence-to-structure mapping, reverse mapping, or bio-activity prediction, making it unrealistic for biologists to utilize such tools. Instead, our work proposes a new method of computer-assisted inhibitor discovery: de novo pocket-aware peptide structure and sequence generation network. Our approach consists of two sequential diffusion models for end-to-end structure generation and sequence prediction. By leveraging angle and dihedral relationships between backbone atoms, we ensure an E(3)-invariant representation of peptide structures. Our results demonstrate that our method achieves comparable performance to state-of-the-art models, highlighting its potential in pocket-aware peptide design. This work offers a new approach for precise drug discovery using receptor-specific peptide generation. The code used for this research is available at Github.\footnote{\url{https://github.com/LabJunBMI/E3-invaraint-diffusion-model-for-pocket-aware-peptide-generation}}

\end{abstract}
\section{Introduction}


Biologists frequently search for peptides to support a range of applications, from understanding biological mechanisms to addressing healthcare challenges. For instance, immunotherapy uses immune checkpoint inhibitors to block checkpoint proteins from binding, thereby increasing the activity of immune cells against cancer cells. Traditionally, discovering peptides that block checkpoint proteins has been a labor-intensive and time-consuming process. However, recent advances in computational methods have significantly accelerated this process \cite{valentinuzzi2020computational,duran2024might}. These computational methods--such as molecular docking, virtual screening, and machine learning/deep learning--have advanced peptide discovery by offering tools for various tasks, including predicting peptide-protein interactions, optimizing peptide sequences, and identifying potential peptide candidates \cite{duran2022molecular,lei2021deep,duran2021molecular}. Among these methods, deep learning is considered one of the most advanced computational approaches. By utilizing neural network architectures with large datasets, these models can identify peptides with desired bioactive properties for various applications. Deep learning methods, including various sequence- or structure-generative models \cite{lin2022novo,liang2024exploring}, have demonstrated success in optimizing peptides and generating peptide structures or sequences with desired properties. This progress has facilitated further research into generating peptides with specific characteristics.

While extensive research has been conducted on computer-aided peptide discovery \cite{sharma2023peptide}, most efforts have focused on generating peptide sequences with general properties \cite{gupta2019feedback,greener2018design} or desired structures \cite{goverde2023novo,goverde2024computational}, which makes these tools less practical for biologists. Only few research \cite{wan2022deep} has been conducted on computer-aided peptide inhibitor discovery. The limitation of general peptide discovery without desired target information and properties can hinder the discovery of peptide inhibitor for particular receptor pocket. 
The RFdiffusion model \cite{watson2022broadly} offers a potential approach by generating peptide structures around target protein pocket residues, followed by using inverse folding models like ESM-IF \cite{hsu2022learning} or ProteinMPNN \cite{dauparas2022robust} to predict possible sequences for those structures. 
However, the limitations of the research are as follows. First, the inverse folding models lack information about pocket residues, which increases the risk of generating peptides with high affinity for various protein structures rather than a specific target. Second, the RFdiffusion model and other inverse folding models are trained on broad datasets. These datasets often include protein complexes with long amino acid sequences, which limits their specificity for the peptide inhibitor generation task.

Our study addresses the challenges associated with generating target protein pocket-aware peptides. We propose a method to generate a target protein pocket favored peptide backbone structures and sequences that take receptor pocket information into account. Our method first generates peptide structures based on the shape of the pocket, followed by sequence generation that incorporates both the peptide structure and comprehensive pocket information. We evaluate our approach against peptide generation methods that combine state-of-the-art structure prediction and sequence prediction models.

\section{Method}
\begin{figure}[htbp]
    \centering
    \includegraphics[width=\columnwidth]{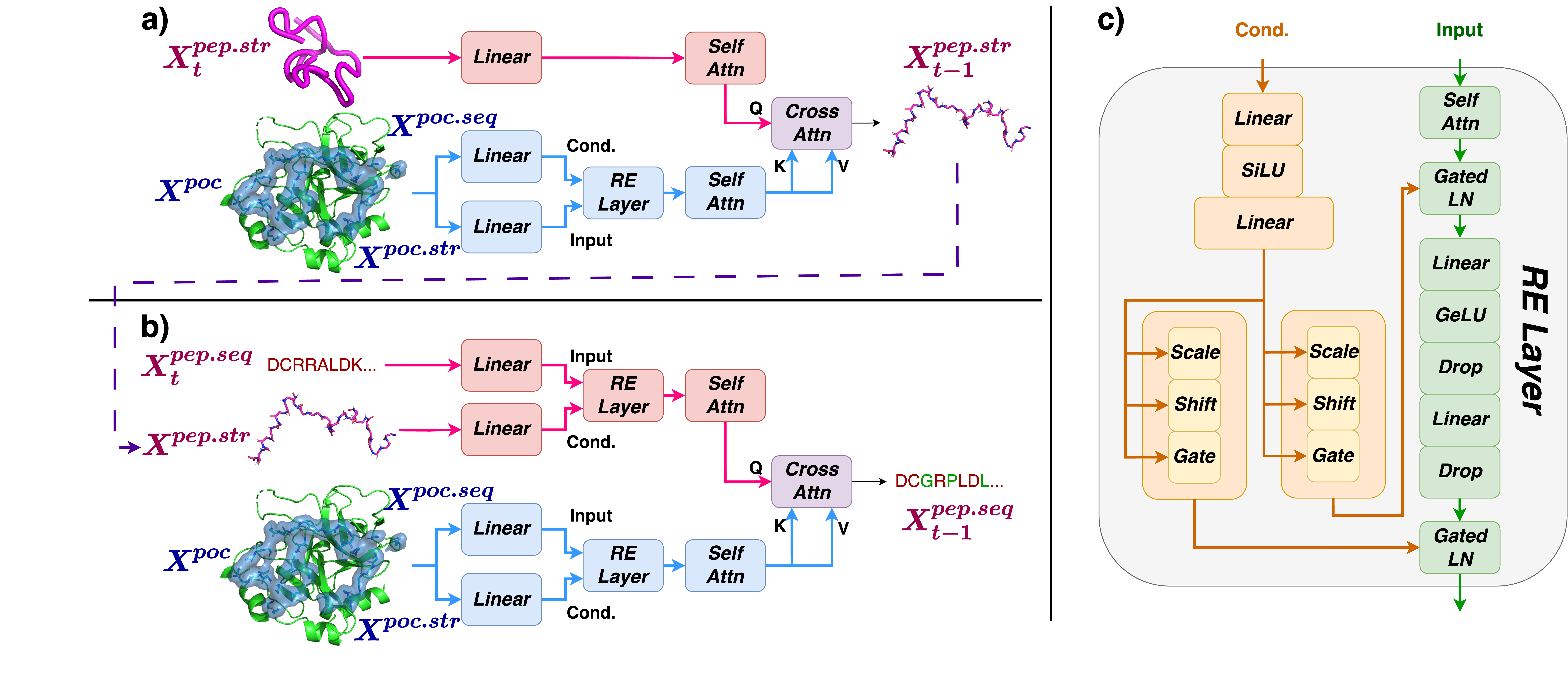}
    \caption{This figure illustrates the main architecture of our method, which includes: a) the architecture of the peptide structure prediction model, b) the architecture of the peptide sequence prediction model, and c) the architecture of a residue encoder (RE) layer. In Figures a and b, the "Self Attn" and "Cross Attn" blocks represent the self-attention and cross-attention mechanisms as proposed in the transformer research \cite{vaswani2017attention}. In Figure c, "input" represents the input hidden state to be normalized, "Cond." denotes the condition, and "gate LN" refers to the gated layer normalization, which is defined by Equation \ref{eq:gate_re}}
    \label{fig:model_structure}
    \vspace{-0.2in}
\end{figure}

Our proposed method (shown in Figure ~\ref{fig:model_structure}) contains twin modules
: 1) the conditional structure diffusion model and 2) conditional sequence diffusion model. The structure diffusion model generate desired peptide structures by given pocket information as condition. Once the peptide structure is obtained, the sequence diffusion model generates the corresponding amino acid sequence, using the pocket information as a condition as well. In this section, we explain the E(3)-invariant  representation for proteins and peptides, as well as the design of our twin conditional diffusion models.

\subsection{Peptide and Pocket Representation}
\label{sec:structure_rep}
\subsubsection{Representation of Peptide} In FoldingDiff \cite{wu2024protein} the backbone structure of an amino acid can be described as a series of bond angles, dihedral (torsion) angles, and bond distances between the backbone nitrogen, $\alpha$-carbon, and carbon atoms. This provides an E(3)-invariant representation of protein structures, which means that the structure remains consistent regardless of rotations or translations in 3D space. 
Inspired by their work, we represent peptide structures using the bond angles and dihedral angles of the complete backbone atoms, with particular emphasis on the inclusion of the oxygen atom. This inclusion is crucial, as demonstrated by secondary structure prediction tools such as DSSP\cite{kabsch1983dictionary}, which utilize information about backbone oxygen atoms to enhance their predictions, leading to a more comprehensive representation of the amino acid structure. The first and last residues are excluded, as we cannot calculate all the internal angles for these residues. The specific definitions of the angles and dihedrals used are provided in Table \ref{tab:angle_dihedral_def}, where each backbone atom is represented as ${AtomType}_{i}$, corresponding to the backbone atom of the \(i\)-th amino acid residue.
We define the representation of a peptide to be $X^{pep} = \{X^{pep.str}, X^{pep.seq}\}$. The matrix $X^{pep.str} \in [-\pi, \pi]^{n\times 8}$ represent the structure of a peptide by 4 bond angles and 4 dihedrals, where $n$ is the length of the peptide. The matrix $X^{pep.seq} \in \mathbb{R}^{n \times 20}$ denotes the one-hot encoded 20 amino acid types.

\begin{table}[h]
\centering
\caption{Amino Acid Backbone angles to represent the structure. To reconstruct a peptide structure, we first calculate the position of the nitrogen atom using $\psi_{i}$ and $\theta_{1}$, followed by the position of the alpha-carbon using $\omega_{i}$ and $\theta_{2}$, then the carbon using $\phi_{i}$ and $\theta_{3}$, and finally the oxygen using $\delta_{i}$ and $\theta_{4}$.
}
\begin{tabular}{|c|l|}
    \hline
    Angle &  Description\\
    \hline
    $\psi_{i}$ & Dihedral torsion of {\itshape $N_{i-1}-C\alpha_{i-1}-C_{i-1}-N_{i}$}\\
    $\omega_{i}$ & Dihedral torsion of {\itshape $C\alpha_{i-1}-C_{i-1}-N_{i}-C\alpha_{i}$}\\
    $\phi_{i}$ & Dihedral torsion of {\itshape $C_{i-1}-N_{i}-C\alpha_{i}-C_{i}$}\\
    $\delta_{i}$ & Dihedral torsion of {\itshape $N_{i}-C\alpha_{i}-C_{i}-O_{i}$}\\
    \hline
    $\theta_{1}$ & Bond angle of {\itshape $C\alpha_{i-1}-C_{i-1}-N_{i}$}\\
    $\theta_{2}$ & Bond angle of {\itshape $C_{i-1}-N_{i}-C\alpha_{i}$}\\
    $\theta_{3}$ & Bond angle of {\itshape $N_{i}-C\alpha_{i}-C_{i}$}\\
    $\theta_{4}$ & Bond angle of {\itshape $C\alpha_{i}-C_{i}-O_{i}$}\\
    \hline
\end{tabular}
\label{tab:angle_dihedral_def}
\end{table}

\subsubsection{Representation of Pocket} \label{pocket_representation}
To create a pocket-aware representation, we describe the receptor pocket as $X^{poc} = \{X^{poc.str}, X^{poc.seq}\}$, where $X^{poc.str} \in [-\pi, \pi]^{m\times 8}$ denotes the structure of each pocket residue using the same structure representation as peptide, and $m$ is the number of pocket residues. $X^{poc.seq} \in \mathbb{R}^{m \times 20}$ denotes the amino acid types of each pocket residues. Additionally, we believe that nearby residues in the sequence may also contribute to the binding process. Therefore, we include the neighboring residues of each pocket residue within a specified range to $X^{poc}$, enhancing our representation and providing more comprehensive information of the interactions that occur in the receptor pocket. This method helps us combine both local structural features and the impact of neighboring residues on binding dynamics. 


\subsection{Twin Conditional Diffusion Model}

\subsubsection{Structure Diffusion Model}
We designed a transformer \cite{vaswani2017attention}-based diffusion model for generating peptide structures conditioned on the receptor pocket. The diffusion model architecture contains two major processes, the Markov noising process and de-noising process. The Markov noising process corrupts the data by progressively adding noise, to it over a series of time steps, which could be represented as $q(X^{pep.str}_{t} | X^{pep.str}_{t-1})$, where $t$ is the discrete time step in the noising process, indicating the level of noise applied to the data at that particular step. To perform the de-noising process, a model, denotes as $p(X^{pep.str}_{t-1} | X^{pep.str}_{t}, X^{poc})$, is trained to reverse a Markov noising process. To capture the wrapped nature of angle, we utilize the wrapped normal distribution for Markov noising process and the wrapped smooth $L_1$ loss, which can be represent by Equation \ref{eq: structure_loss}, for model training, introduced in FoldingDiff \cite{wu2024protein}.

\begin{align}
    d(\cdot) = ((\epsilon_{t} - \epsilon_{p} +\pi)\mod 2\pi) - \pi \nonumber \\
    L_{str} =
    \begin{cases}
        0.5d(\cdot)/\beta  , & \text{if } |d(\cdot)| < \beta \\
        |d(\cdot)| - 0.5\beta, & \text{otherwise}
    \end{cases}
    \label{eq: structure_loss}
\end{align}
where the function $d(\cdot)$ wrap the angle difference about $[-\pi, \pi)$, $\epsilon_{t}$ represent the noise sampled from noising process, $\epsilon_{p}$ represent the noise predicted by our model, and $\beta$ is a hyper-parameter, which we set to $0.1 \pi$.

\subsubsection{Sequence Diffusion Model}
In order to inverse fold the generated peptide structure, we designed another discrete diffusion model. Follow the work done by GraDe-IF\cite{yi2024graph} research, for the corrupted $n$-th amino acid type of a sequence, which can be represent as $X^{seq}_{t}[n] = A_{t} \in \mathbb{R}^{20}$, the transition probability between amino acid type $i$ and $j$ at step $t$ can be described as a matrix $[Q_{t}]_{ij}=q(A_{t}=i | A_{t-1}=j)$. The noising process could be defined as $q(A_{t} | A_{t-1})=A_{t-1}Q_{t}$. To reverse the noising process, a model $p(X^{pep.seq}_{0} | X^{pep.seq}_{t}, X^{pep.str}, X^{poc})$ is trained to estimate the $p(X^{pep.seq}_{t-1} | x^{pep.seq}_{t}, x^{pep.str}, x^{poc})$. We adapt the BLOSUM transition matrix proposed in GraDe-IF \cite{yi2024graph} to better capture the property of transition between amino acid types. To motivate the model learning the nature distribution of amino acids types in peptides, we construct the loss by combining the ELBO loss and cross entropy, shown as the Equation \ref{eq:structure_loss}.

\begin{equation}
    L_{seq} = L_{ELBO} + L_{CE}
    \label{eq:structure_loss}
\end{equation}

where the $L_{ELBO}$ loss  (shown in Equation ~\ref{equ:elbo}) is applied measure the distribution divergence between the amino acid from ground truth and prediction, and the $L_{CE}$ (shown in Equation ~\ref{eq:seq_loss}) is applied to measure the probability distance between each individual residue amino acid between ground truth and prediction.

\begin{equation}
     L_{ELBO} = \mathbb{E}_{q(\hat{A}|A_{0})} [ \log p(A_{0}|\hat{A}) ] - KL( q(\hat{A}|A_{0}) \| p(\hat{A}) ) 
     \label{equ:elbo}
\end{equation}
where $KL$ is the Kullback–Leibler divergence, and $\hat{A}$ represent the prediction from model.
\begin{align}  
    L_{CE} = - \sum_{i}^{20}p(i) \log q(i) 
    \label{eq:seq_loss}
\end{align}
where functions $p$ and $q$ represent the ground truth and predicted probability distribution respectively.
\subsection{Conditional Layers}
\subsubsection{Residue Encoding(RE) Layer}
To encode the angle and residue type features of residues, we employ gated adaptive layer normalization \cite{huang2017arbitrary,liu2024inverse}, as illustrated in Figure \ref{fig:model_structure}c. In the structural diffusion model, we consider residue type as a conditioning variable for the structural features, whereas in the sequence diffusion model, the relationship is reversed. This approach allows us to maintain the essential characteristics of one type of feature while also integrating the effects of the other.

\begin{equation}
    GatedLN(\cdot) = Gate \odot (Scale \odot \frac{h-\mu(h)}{\sigma(h)}+Shift)
    \label{eq:gate_re}
\end{equation}
where $h$ is the hidden state to be normalized, $Gate$, $Scale$, and $Shift$ are the layer normalization parameters derived from condition hidden state, and the $\odot$ operation represent the Hadamard product.

\subsubsection{Cross-Attention Layer}

After obtaining the embeddings for the pocket and peptide, represented as and  respectively, we condition the peptide features using the pocket features through a cross-attention mechanism \cite{vaswani2017attention}. In this mechanism, the query $Q$ is calculated as $W_{q}h^{pep}$, while the key $K$ and value $V$ are derived from the pocket features using $W_{k}h^{pocket}$ and $W_{v}h^{pocket}$. The output of this layer can be expressed as $h_{out} = Softmax(QK^T/\sqrt{d_K}) \cdot V$ where $d_k$ denotes the dimension of $K$. The resulting output represents the conditioned features, effectively integrating information from both the pocket and peptide embeddings.

\subsection{Data Source and Evaluation Metrics}
\subsubsection{Data Source and Preprocessing}
To evaluate the performance of our model, we used the open-source BioLip \cite{zhang2024biolip2} protein-ligand dataset. The Biopython\cite{cock2009biopython} package and the DSSP \cite{kabsch1983dictionary} tool were utilized for data preprocessing. The BioLip database contains 781,684 protein-ligand interactions, including 35,167 interactions specifically between proteins and peptides. To ensure data quality, we excluded complexes with a resolution lower than $5$ Å \cite{salamanca2017optimal} and those containing unknown amino acids. Additionally, we removed peptides shorter than 5 amino acids, complexes with duplicate PDB IDs, and those that could not be processed by Biopython or DSSP. After these filtering steps, we obtained a final set of 8,868 receptor-peptide complexes, which were divided into training, validation, and testing sets with an 80:10:10 ratio.

\subsubsection{Model Testing Procedure}
we tested a version trained on a dataset with four extended (ext-$4$) neighbor residues, as detailed in Section \ref{pocket_representation}. This model, referred to as OurModel, was chosen because it includes most of the relevant pocket residue information. We also tested a combined model that integrates residues from $0$ to $4$ extended neighbors (denoted as OurModel-ext), which represents the best performance by using different ranges of neighboring residues. For overall binding performance, the ext-${i}$ structure model results were used as input for the corresponding ext-${i}$ sequence model, ensuring alignment of pocket residue information.

Among the models we compared against, only the diffusion-based models, RFdiffusion \cite{watson2022broadly} and GraDe-IF \cite{yi2024graph}, have the inherent ability to generate varied results from the same input. To ensure a fair comparison, the other models were tested using a one-shot generation approach. Additionally, performing multiple generations for all these models is impractical, as it would exponentially increase the number of structure-sequence pairs and significantly extend the required post-processing time.

\subsubsection{Structure Evaluation Metrics}
We evaluate the performance of our peptide structure prediction model against RFdiffusion \cite{watson2022broadly} using Root Mean Square Deviation (RMSD) of backbone atoms and TM-Score, a length-normalized structure similarity metric. Given the limited research in this domain, we also benchmark our model against other folding models, including AlphaFold2\cite{jumper2021highly}, ESM Fold\cite{lin2023evolutionary}, and OmegaFold \cite{wu2022high}, which predict structures based on amino acid sequences, providing a contrast to non-pocket-aware models. To ensure a fair comparison, we assess our model's performance using a reconstructed peptide structure, where bond distances are fixed for each bond type, while other models are evaluated against the original peptide structures. For AlphaFold2, we utilized the ColabFold \cite{mirdita2022colabfold,Mirdita2017,Mitchell2019} implementation, which employs MMseqs2 server \cite{mirdita2019mmseqs2} to reduce the time required for structure prediction. Five pre-trained AlphaFold2 models, each trained with different random seeds, are available. We used all five and selected the structure with the highest confidence score (pLDDT). In the FoldingDiff research \cite{wu2024protein}, they demonstrated the ability of their model to capture overall angle distributions and secondary structures. To evaluate whether these properties are maintained after incorporating the pocket information into the generation process, we also assess the divergence of angles between the testing set and the generated set to determine whether our model can generate structures with a similar angle distribution. Additionally, the Ramachandran plot is used to visualize our model's ability to capture secondary structure properties.
\subsubsection{Sequence Evaluation Metrics}\label{seq_evaluation_method}
The performance of the sequence generation model is evaluated using the commonly used amino acid sequence recovery rate. Nonetheless, the recovery rate may underestimate performance when the model has a shifted prediction. For example, if the predicted sequence has one amino acid inserted at the start of a sequence, the recovery rate will drop to 0. Therefore, we propose a global alignment score-based sequence similarity evaluation, which can be described by Equation \ref{eq:seq_similarity}, where $N.W.$ represents the Needleman-Wunsch algorithm \cite{needleman1970general} used to calculate the global alignment score. The numerator term evaluates the similarity of two sequences, while the denominator normalizes the score by the maximum possible alignment score, known as the self-alignment score. Following a similar concept, we also designed a diversity metric defined by Equation \ref{eq:seq_diversity}, where $X^{seq}_i$ represents the $i$-th sequence within a sequence set $X^{seq}$. The diversity metric is defined as one minus the average normalized pairwise alignment scores. If the sequence set has high diversity, meaning the sequences are different from each other, the pairwise alignment scores will be low, resulting in a high diversity score, and vice versa. One of the advantages of this method is that we can incorporate a substitution matrix to evaluate diversity, allowing for the inclusion of natural mutation information. The substitution matrix can also be defined to exclude any mutation information, providing the user with flexibility to adjust the metric to better fit specific problems. In our research, the most commonly used BLOSUM62 \cite{henikoff1992amino} substitution matrix is used to calculate sequence similarity and diversity.

\begin{equation}
    Sequence Similarity = \frac{N.W.(x_{pred}^{seq}, x_{true}^{seq})}{N.W.(x_{true}^{seq}, x_{true}^{seq})}
    \label{eq:seq_similarity}
\end{equation}

\begin{align} \label{eq:seq_diversity}
    SequenceDiversity = 1-\frac{1}{N}\frac{1}{N} \sum_i^N\sum_j^N \frac{N.W.(X^{seq}_i, X^{seq}_j)}{N.W.(X^{seq}_j, X^{seq}_j)}\\ 
    , \text{where length of $X^{seq}_j$ > length of $X^{seq}_i$} \nonumber
\end{align}

\subsubsection{Docking Evaluation Metrics}
The overall performance of our model is evaluated by estimating the energy between the receptor and the generated peptide using PyRosetta \cite{chaudhury2010pyrosetta,leaver2013scientific}. The generated peptide is aligned with the position of the original peptide, followed by side-chain packing and a fast relaxation step \cite{simons1999improved,renfrew2014rotamer}. We then use the Rosetta energy function to estimate the energy sums for the individual receptor and peptide structures, as well as the energy of the combined complex. The difference between these energy terms represents the overall docking energy of the complex. Additionally, we use the energy from the docking protocol \cite{davis2009rosettaligand} to provide an alternative perspective on model performance. Beyond the estimated energy terms, we also assess whether the peptide maintains contact with the binding pocket after the fast relaxation. A distance cutoff of $5$ Å \cite{salamanca2017optimal} is used to define contact, determined by whether peptide residues are in contact with any pocket residues. This definition aligns with a drug discovery perspective, where a peptide can influence signaling pathways as long as some pocket residues are blocked.
\section{Result}

\subsection{Structure Prediction}

The peptide structure prediction results presented in Table \ref{tab:structure_performance} indicate that the integrated result of our model (Our Model-ext) outperforms all other models across all metrics. In comparison, the non-integrated result of our model has a slightly lower RMSD (by 0.18 Å) than the second-best result from ESM Fold. However, it achieves a higher TM Score than the other models in terms of average score, as well as the ratios of scores exceeding 0.2 and 0.5. This may be attributed to two factors: 1) the inclusion of extra pocket residue information significantly reduces the search space for peptide structures and 2) our model is specifically trained on peptide (short protein) structures, which may exhibit distribution differences compared to general protein datasets. Furthermore, while AlphaFold2 and ESM Fold perform better in RMSD, they have lower TM Scores (which are normalized by length), reinforcing this hypothesis.

\begin{table}
\centering
\caption{Peptide Structure prediction performance comparing to other models. The highest 1\% of RMSD data points was removed to mitigate the influence of extreme outliers.}
\vspace{-0.1in}
\label{tab:structure_performance}
\begin{tabular}{c c| c @{\hspace{1em}} c @{\hspace{1em}} c | c @{\hspace{1em}} c @{\hspace{1em}} c}
& & \multicolumn{3}{c}{\textbf{Non Pocket-Aware}} & \multicolumn{3}{|c}{\textbf{Pocket-Aware}} \\
&& AlphaFold2 & ESM-Fold & OmegaFold & RFDiffusion & OurModel & OurModel-ext\\
\hline
$\downarrow$ &Average RMSD(Å)\rule{0pt}{2.5ex}  & 3.82 & \underline{3.28} & 7.41 & 4.54 & 3.46 & \textbf{2.49} \\
$\uparrow$ &RMSD<5Å & 78.71\% & \underline{80.18\%} & 61.37\% & 65.87\% & 75.90\% & \textbf{90.31\%} \\
\hline
$\uparrow$&Average TM-Score & 0.14 & 0.14 & 0.11 & 0.15 & \underline{0.31} & \textbf{0.42} \\
$\uparrow$&TM-Score>0.2 & 32.77\% & 31.53\% & 25.90\% & 29.96\% & \underline{65.65\%} & \textbf{89.86\%} \\
$\uparrow$&TM-Score>0.5 & 4.94\% & 3.60\% & 2.25\% & 6.86\% & \underline{15.09\%} & \textbf{29.39\%} \\
\hline
\multicolumn{8}{l}{\textit{Best score is in \textbf{bold}; second-best score is \underline{underlined}.}}
\end{tabular}
\end{table}

\begin{figure}
    \centering
    \includegraphics[height=5.7cm]{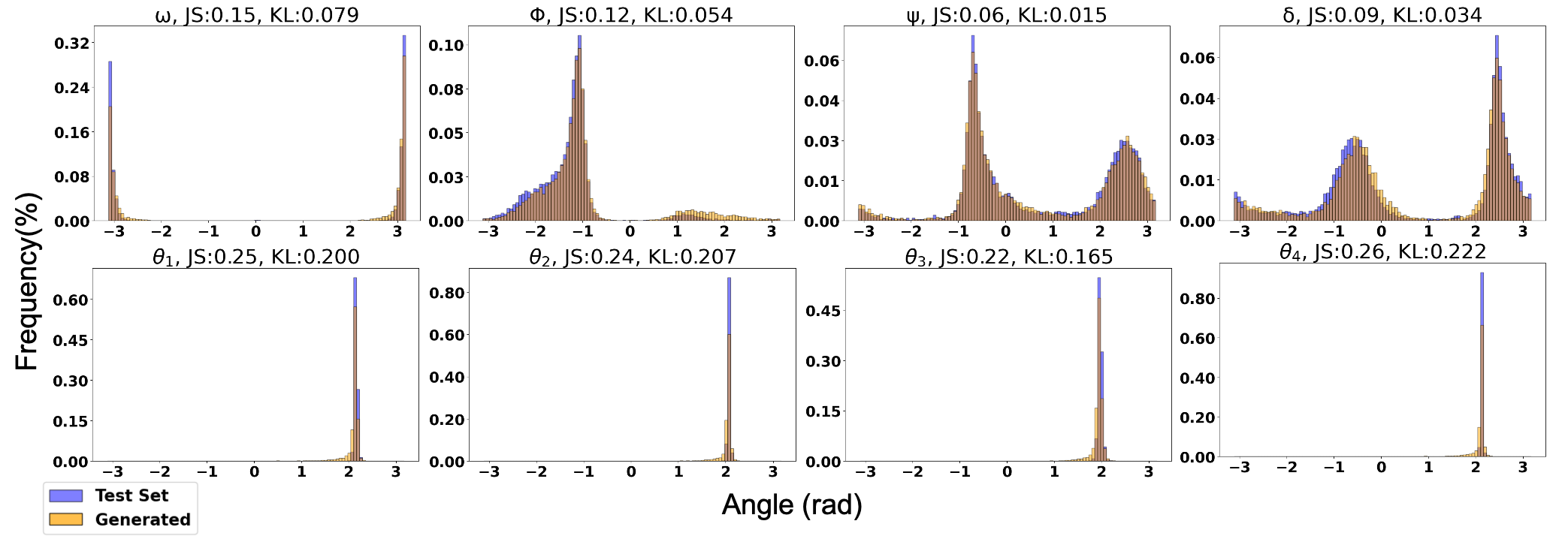}
    \caption{This figure compares the distributions of generated angles with those from testing set. The Jensen-Shannon (JS) distance and Kullback-Leibler (KL) divergence are used to quantify the differences between two distributions. The top four plots illustrates the dihedral angle distributions, while the bottom plots displays the bond angle distributions.}
    \label{fig:angle_distribution}
\end{figure}

In addition to RMSD and TM-Score, we examined whether the generated structures exhibit similar angle distributions to those in the testing set. As shown in Figure \ref{fig:angle_distribution}, the dihedral angle distributions are highly comparable. While the $\omega$ angle demonstrates a slightly greater divergence, it still maintains a strong similarity  to the testing set. In contrast, the bond angle distributions show greater divergence compared to the dihedrals, primarily due to their narrower range of distribution. Nevertheless, the model effectively captures the range of bond angles, with the majority of the generated bond angles falling within the distribution range of the testing set. 
\begin{figure}

    \centering
    \includegraphics[width=0.5\linewidth]{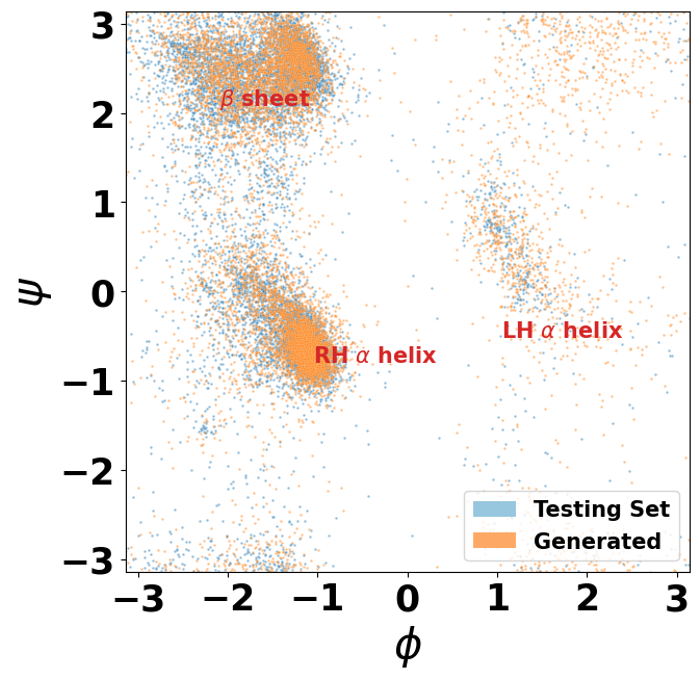}
    \caption{Ramachandran plot visualizes the distribution of secondary structure by comparing the $\phi$ and $\psi$ dihedral angles of amino acids. The areas corresponding to three main secondary structures -- namely $\beta$ sheet, right-handed (RH) $\alpha$ helix, and left-handed (LH) $\alpha$ helix -- are marked on the plot.}
    \label{fig:Ramachandran}
\end{figure}

Figure \ref{fig:Ramachandran} presents a Ramachandran plot comparing the structures from the testing set with those generated by our model. This figure shows that our model effectively captures and distinguishes between the three main secondary structures: the $\beta$ sheet, right-handed (RH) $\alpha$ helix, and left-handed (LH) $\alpha$ helix. Notably, the LH $\alpha$ helix is more challenging to capture due to its lower prevalence. This illustrates that, even with conditioning incorporated into the generation process, the model's ability to accurately represent secondary structures in this angle-based structure generation remains robust.

\subsection{Sequence Prediction}
To evaluate the sequence prediction performance, we predicted sequences based on the original peptide structure and compared the results with three state-of-the-art inverse folding models: ESM-IF \cite{hsu2022learning}, ProteinMPNN \cite{dauparas2022robust}, and GraDe-IF \cite{yi2024graph}. The results shows in Table \ref{tab:sequence_performance}, indicate that our model significantly outperforms these models, achieving a recovery rate of 47.41\% with non-integrated prediction results, where recovery rate refers to the percentage of correctly predicted amino acids in the sequence. Additionally, the diversity of the generated sequences, as defined in section \ref{seq_evaluation_method}, is higher compared to the other three models. The diversity between non-integrated and integrated predictions is essentially the same, as this metric measures variation within a set from the same generation round, and the diversity within a single model remains relatively stable across different repetitions. 

\begin{table}
\centering
\caption{Peptide Sequence prediction performance comparing to other models. Details of $Sequence Similarity$ and $Diversity$ are defined in \ref{seq_evaluation_method}}\label{tab:sequence_performance}
\begin{tabular}{c c| c @{\hspace{1em}} c @{\hspace{1em}} c @{\hspace{1em}} | c @{\hspace{1em}} c}
& & ESM-IF & ProteinMPNN & GraDe-IF & OurModel & OurModel-ext\\
\hline
$\uparrow$ & Recovery Rate  & 17.86\% & 19.04\% & 19.92\% & \underline{47.41\%} & \textbf{59.76\%} \\
$\uparrow$ & Sequence Similarity & 0.34 & 0.33 & 0.35 & \underline{0.59} & \textbf{0.70} \\
$\uparrow$ & Diversity & 0.57 & 0.57 & 0.60 & \underline{0.64} & \underline{0.64} \\
\hline
\multicolumn{7}{l}{\textit{Best score is in \textbf{bold}; second-best score is \underline{underlined}.}}
\end{tabular}

\end{table}

A key factor contributing to the notable performance gap is the incorporation of pocket residue information in our proposed method. This information could help the model to reduces the search space for potential amino acid types, enabling more accurate sequence predictions. Furthermore, differences in data distribution between the training and testing sets, such as peptide length and structure difference, also play a role. Our model is specifically trained on peptides, which are short protein sequences, all fewer than 30 amino acids in our dataset. In contrast, the other models are trained on more general protein datasets that include sequences with hundreds of amino acids. This further underscores the importance of our peptide-specific model in accurately predicting sequences for short peptides.

\subsection{Binding Analysis}
We evaluate binding effectiveness using three metrics: binding energy, docking score, and contact rate. Table \ref{tab:all_binding_energy} presents the results for all combinations of structure generation and sequence generation models. When comparing different structure generation models, the RFdiffusion model outperforms the others in terms of binding energy and contact rate across all sequence generation methods, despite lacking information about the actual structure of the peptide. In terms of dock score, the OmegaFold model performs slightly better (by less than 5\%). Table \ref{tab:our_binding_energy} shows that the integrated result of our model are comparable to those of RFdiffusion and surpass other non-pocket-aware methods in terms of binding energy and contact rate. It is important to note that while the pocket-aware method achieves higher binding energy, its docking score is lower compared to non-pocket-aware methods. This discrepancy may be attributed to the evaluation method used for the docking score, which measures the energy of the entire peptide-receptor complex. If the energy of a peptide is higher in one structure compared to another, the overall energy of the complex will also be higher.This indicates that the energy difference may reflect the characteristics of peptides rather than the interaction energy between the peptide and the receptor.
\begin{table}
\centering
\caption{Docking performance of peptides generated by combining the folding model and inverse folding model. Dock score denotes the energy obtained from dock protocol, and Cont. Rate is the contact rate, represent the percentage of peptide has interaction with pocket residues.\textbf{ For the sequence-shuffled (Seq. Shuffled) results, the values in parentheses represent the difference between shuffled and unshuffled results. A larger difference indicates better performance.}}\label{tab:all_binding_energy}
\begin{tabular}{c c |c| c@{\hspace{1em}} c@{\hspace{1em}} c | c | c @{\hspace{1em}} c }
&\multicolumn{2}{c|}{} & \multicolumn{3}{c}{\textbf{Non Pocket-Aware}} & \multicolumn{1}{|c}{\textbf{Pocket-Aware}} & \multicolumn{2}{|c}{\textbf{Seq. Shuffled}}\\
&\multicolumn{2}{c|}{} & AlphaFold2 & ESM-Fold & OmegaFold & RFdiffusion & AlphaFold2 & RFdiffusion \\
\hline
&& ESM-IF & -28.95 & -28.27 &-29.76&\underline{-32.70}&-28.85 (\underline{0.10})&-32.11 (\textbf{0.59})\\
$\uparrow$&Binding Energy & ProteinMPNN & -29.49 & -29.60 &-30.05&-32.24&-29.49 (0.00)&-32.24 (0.00)\\
&& GraDe-IF & -30.61 & -30.88 &-31.61&\textbf{-32.88}&-30.54 (0.07)&-32.78 (\underline{0.10})\\
\hline

&& ESM-IF & -750.19 & -780.82 &\underline{-784.95}&-747.94&-757.99 (7.8)&-742.29 (5.65)\\
$\uparrow$&Dock Score & ProteinMPNN & -752.25 &-779.64 &\textbf{-790.22}&-748.79&-752.26 (0.01)&-748.80 (0.01)\\
&& GraDe-IF & -737.32 & -770.83 &-772.06&-751.33&-774.80 (\textbf{37.48})&-742.89 (\underline{8.44})\\
\hline

&& ESM-IF & 96.39\% & 96.73\% &89.07\%&98.31\%&96.50\% (0.11\%)&98.54\% (\underline{0.23\%})\\
$\uparrow$&Cont. Rate & ProteinMPNN & 96.50\% & 97.63\% &88.29\%&\underline{98.87\%}&96.50\% (0.00\%)&98.87\% (0.00\%)\\
&& GraDe-IF & 96.14\% &97.52\% &88.85\%&\textbf{99.21\%}&96.05\% (0.09\%)&98.09\% (\textbf{1.12\%})\\
\hline
\multicolumn{9}{l}{\textit{Best score is in \textbf{bold}; second-best score is \underline{underlined}.}}
\end{tabular}
\end{table}

Beyond the straight forward binding performance, we also assessed the specificity of the generated peptide sequences by shuffling the sequences while maintaining the same length. Our hypothesis is that if a peptide has high specificity, its performance should decrease when tested against other pockets. We tested the widely used non-pocket-aware model, AlphaFold2, and the pocket-aware model, RFdiffusion. The rightmost two columns in Table \ref{tab:all_binding_energy} show the results with shuffled sequences, and surprisingly, the performance difference between the shuffled and unshuffled results is negligible, remaining nearly unchanged across all combinations and metrics.In contrast, when our model is tested with shuffled sequences, as shown in Table \ref{tab:our_binding_energy}, there is a larger drop in performance, achieving only 40.59\% of the original performance in binding energy. No decrease in docking score was observed, likely because the docking score measures the overall energy of the entire complex, with the value being dominated by the larger receptor structure compared to the smaller peptide. This result indicates that our model is better at generating peptides with higher specificity while maintaining comparable performance.

\begin{table}
\centering
\caption{Docking performance of peptides generated by our model compared to peptides from the testing set. \textbf{ For the sequence-shuffled (OurModel-Seq. Shuffled) results, the values in parentheses represent the difference between shuffled and unshuffled results. A larger difference indicates better performance.}}
\label{tab:our_binding_energy}
\begin{tabular}{c c | c | c @{\hspace{1em}} c |c}
& & Testing Set & OurModel & OurModel-ext & OurModel-Seq. Shuffled\\
\hline
$\uparrow$ & Binding Energy&-46.16&-20.32&-31.62&-8.58 (11.74)\\
\hline
$\uparrow$ & Dock Score&-789.02& -489.95&-698.34&-504.27 (14.32)\\
\hline
$\uparrow$ & Cont. Rate&99.09\%&89.63\%&98.42\%&89.88\% (0.25\%)\\
\hline
\multicolumn{6}{l}{\textit{Best score is in \textbf{bold}; second-best score is \underline{underlined}.}}
\end{tabular}
\end{table}

\section{Discussion}

In this research, we proposed a new method for generating pocket-aware peptides using diffusion models integrated with E(3)-invariant structure representation. The results show that our model can predict peptide structures with low RMSD, indicating high accuracy. Additionally, the TM-Score results further demonstrate that our model performs better in generating accurate peptide structures compared to both pocket-aware and non-pocket-aware models. Our approach also achieves a higher recovery rate due to the inclusion of extra pocket information. Moreover, the increased sequence diversity suggests that our model can design peptide sequences with greater specificity for the given receptor pocket residues, rather than merely generating sequences with high affinity for any receptor. This is further supported by the shuffled sequence binding test results, where our model exhibited a drop in binding energy, while the others remained largely unchanged. In conclusion, this research emphasizes the importance of pocket residue information in advancing peptide design, highlighting our model's potential for applications such as immune checkpoint inhibitors.
\newpage
%
%
%
\bibliographystyle{splncs04}
\bibliography{ref}

\begin{thebibliography}{10}
\providecommand{\url}[1]{\texttt{#1}}
\providecommand{\urlprefix}{URL }
\providecommand{\doi}[1]{https://doi.org/#1}

\bibitem{chaudhury2010pyrosetta}
Chaudhury, S., Lyskov, S., Gray, J.J.: Pyrosetta: a script-based interface for implementing molecular modeling algorithms using rosetta. Bioinformatics  \textbf{26}(5),  689--691 (2010)

\bibitem{cock2009biopython}
Cock, P.J., Antao, T., Chang, J.T., Chapman, B.A., Cox, C.J., Dalke, A., Friedberg, I., Hamelryck, T., Kauff, F., Wilczynski, B., et~al.: Biopython: freely available python tools for computational molecular biology and bioinformatics. Bioinformatics  \textbf{25}(11),  1422--1423 (2009)

\bibitem{dauparas2022robust}
Dauparas, J., Anishchenko, I., Bennett, N., Bai, H., Ragotte, R.J., Milles, L.F., Wicky, B.I., Courbet, A., de~Haas, R.J., Bethel, N., et~al.: Robust deep learning--based protein sequence design using proteinmpnn. Science  \textbf{378}(6615),  49--56 (2022)

\bibitem{davis2009rosettaligand}
Davis, I.W., Baker, D.: Rosettaligand docking with full ligand and receptor flexibility. Journal of molecular biology  \textbf{385}(2),  381--392 (2009)

\bibitem{duran2024might}
Duran, T., Chaudhuri, B.: Where might artificial intelligence be going in pharmaceutical development? (2024)

\bibitem{duran2021molecular}
Duran, T., Minatovicz, B., Bai, J., Shin, D., Mohammadiarani, H., Chaudhuri, B.: Molecular dynamics simulation to uncover the mechanisms of protein instability during freezing. Journal of Pharmaceutical Sciences  \textbf{110}(6),  2457--2471 (2021)

\bibitem{duran2022molecular}
Duran, T., Minatovicz, B., Bellucci, R., Bai, J., Chaudhuri, B.: Molecular dynamics modeling based investigation of the effect of freezing rate on lysozyme stability. Pharmaceutical Research  \textbf{39}(10),  2585--2596 (2022)

\bibitem{goverde2024computational}
Goverde, C.A., Pacesa, M., Goldbach, N., Dornfeld, L.J., Balbi, P.E., Georgeon, S., Rosset, S., Kapoor, S., Choudhury, J., Dauparas, J., et~al.: Computational design of soluble and functional membrane protein analogues. Nature pp. 1--10 (2024)

\bibitem{goverde2023novo}
Goverde, C.A., Wolf, B., Khakzad, H., Rosset, S., Correia, B.E.: De novo protein design by inversion of the alphafold structure prediction network. Protein Science  \textbf{32}(6),  e4653 (2023)

\bibitem{greener2018design}
Greener, J.G., Moffat, L., Jones, D.T.: Design of metalloproteins and novel protein folds using variational autoencoders. Scientific reports  \textbf{8}(1),  16189 (2018)

\bibitem{gupta2019feedback}
Gupta, A., Zou, J.: Feedback gan for dna optimizes protein functions. Nature Machine Intelligence  \textbf{1}(2),  105--111 (2019)

\bibitem{henikoff1992amino}
Henikoff, S., Henikoff, J.G.: Amino acid substitution matrices from protein blocks. Proceedings of the National Academy of Sciences  \textbf{89}(22),  10915--10919 (1992)

\bibitem{hsu2022learning}
Hsu, C., Verkuil, R., Liu, J., Lin, Z., Hie, B., Sercu, T., Lerer, A., Rives, A.: Learning inverse folding from millions of predicted structures. biorxiv (2022). preprint  (2022)

\bibitem{huang2017arbitrary}
Huang, X., Belongie, S.: Arbitrary style transfer in real-time with adaptive instance normalization. In: Proceedings of the IEEE international conference on computer vision. pp. 1501--1510 (2017)

\bibitem{jumper2021highly}
Jumper, J., Evans, R., Pritzel, A., Green, T., Figurnov, M., Ronneberger, O., Tunyasuvunakool, K., Bates, R., {\v{Z}}{\'\i}dek, A., Potapenko, A., et~al.: Highly accurate protein structure prediction with alphafold. nature  \textbf{596}(7873),  583--589 (2021)

\bibitem{kabsch1983dictionary}
Kabsch, W., Sander, C.: Dictionary of protein secondary structure: pattern recognition of hydrogen-bonded and geometrical features. Biopolymers: Original Research on Biomolecules  \textbf{22}(12),  2577--2637 (1983)

\bibitem{leaver2013scientific}
Leaver-Fay, A., O'meara, M.J., Tyka, M., Jacak, R., Song, Y., Kellogg, E.H., Thompson, J., Davis, I.W., Pache, R.A., Lyskov, S., et~al.: Scientific benchmarks for guiding macromolecular energy function improvement. In: Methods in enzymology, vol.~523, pp. 109--143. Elsevier (2013)

\bibitem{lei2021deep}
Lei, Y., Li, S., Liu, Z., Wan, F., Tian, T., Li, S., Zhao, D., Zeng, J.: A deep-learning framework for multi-level peptide--protein interaction prediction. Nature communications  \textbf{12}(1), ~5465 (2021)

\bibitem{liang2024exploring}
Liang, P.Y., Huang, X., Duran, T., Wiemer, A.J., Bai, J.: Exploring latent space for generating peptide analogs using protein language models. arXiv preprint arXiv:2408.08341  (2024)

\bibitem{lin2022novo}
Lin, E., Lin, C.H., Lane, H.Y.: De novo peptide and protein design using generative adversarial networks: an update. Journal of Chemical Information and Modeling  \textbf{62}(4),  761--774 (2022)

\bibitem{lin2023evolutionary}
Lin, Z., Akin, H., Rao, R., Hie, B., Zhu, Z., Lu, W., Smetanin, N., Verkuil, R., Kabeli, O., Shmueli, Y., et~al.: Evolutionary-scale prediction of atomic-level protein structure with a language model. Science  \textbf{379}(6637),  1123--1130 (2023)

\bibitem{liu2024inverse}
Liu, G., Xu, J., Luo, T., Jiang, M.: Inverse molecular design with multi-conditional diffusion guidance. arXiv preprint arXiv:2401.13858  (2024)

\bibitem{Mirdita2017}
Mirdita, M., von~den Driesch, L., Galiez, C., Martin, M.J., S{"{o}}ding, J., Steinegger, M.: {Uniclust databases of clustered and deeply annotated protein sequences and alignments}. Nucleic Acids Res.  \textbf{45}(D1),  D170--D176 (2017). \doi{10.1093/nar/gkw1081}

\bibitem{mirdita2022colabfold}
Mirdita, M., Sch{\"u}tze, K., Moriwaki, Y., Heo, L., Ovchinnikov, S., Steinegger, M.: Colabfold: making protein folding accessible to all. Nature methods  \textbf{19}(6),  679--682 (2022)

\bibitem{mirdita2019mmseqs2}
Mirdita, M., Steinegger, M., S{\"o}ding, J.: Mmseqs2 desktop and local web server app for fast, interactive sequence searches. Bioinformatics  \textbf{35}(16),  2856--2858 (2019)

\bibitem{Mitchell2019}
Mitchell, A.L., Almeida, A., Beracochea, M., Boland, M., Burgin, J., Cochrane, G., Crusoe, M.R., Kale, V., Potter, S.C., Richardson, L.J., Sakharova, E., Scheremetjew, M., Korobeynikov, A., Shlemov, A., Kunyavskaya, O., Lapidus, A., Finn, R.D.: {MGnify: the microbiome analysis resource in 2020}. Nucleic Acids Res.  (2019). \doi{10.1093/nar/gkz1035}

\bibitem{needleman1970general}
Needleman, S.B., Wunsch, C.D.: A general method applicable to the search for similarities in the amino acid sequence of two proteins. Journal of molecular biology  \textbf{48}(3),  443--453 (1970)

\bibitem{renfrew2014rotamer}
Renfrew, P.D., Craven, T.W., Butterfoss, G.L., Kirshenbaum, K., Bonneau, R.: A rotamer library to enable modeling and design of peptoid foldamers. Journal of the American Chemical Society  \textbf{136}(24),  8772--8782 (2014)

\bibitem{salamanca2017optimal}
Salamanca~Viloria, J., Allega, M.F., Lambrughi, M., Papaleo, E.: An optimal distance cutoff for contact-based protein structure networks using side-chain centers of mass. Scientific reports  \textbf{7}(1), ~2838 (2017)

\bibitem{sharma2023peptide}
Sharma, K., Sharma, K.K., Sharma, A., Jain, R.: Peptide-based drug discovery: Current status and recent advances. Drug Discovery Today  \textbf{28}(2),  103464 (2023)

\bibitem{simons1999improved}
Simons, K.T., Ruczinski, I., Kooperberg, C., Fox, B.A., Bystroff, C., Baker, D.: Improved recognition of native-like protein structures using a combination of sequence-dependent and sequence-independent features of proteins. Proteins: Structure, Function, and Bioinformatics  \textbf{34}(1),  82--95 (1999)

\bibitem{valentinuzzi2020computational}
Valentinuzzi, D., Jeraj, R.: Computational modelling of modern cancer immunotherapy. Physics in Medicine \& Biology  \textbf{65}(24),  24TR01 (2020)

\bibitem{vaswani2017attention}
Vaswani, A.: Attention is all you need. Advances in Neural Information Processing Systems  (2017)

\bibitem{wan2022deep}
Wan, F., Kontogiorgos-Heintz, D., de~la Fuente-Nunez, C.: Deep generative models for peptide design. Digital Discovery  \textbf{1}(3),  195--208 (2022)

\bibitem{watson2022broadly}
Watson, J.L., Juergens, D., Bennett, N.R., Trippe, B.L., Yim, J., Eisenach, H.E., Ahern, W., Borst, A.J., Ragotte, R.J., Milles, L.F., et~al.: Broadly applicable and accurate protein design by integrating structure prediction networks and diffusion generative models. BioRxiv pp. 2022--12 (2022)

\bibitem{wu2024protein}
Wu, K.E., Yang, K.K., van~den Berg, R., Alamdari, S., Zou, J.Y., Lu, A.X., Amini, A.P.: Protein structure generation via folding diffusion. Nature communications  \textbf{15}(1), ~1059 (2024)

\bibitem{wu2022high}
Wu, R., Ding, F., Wang, R., Shen, R., Zhang, X., Luo, S., Su, C., Wu, Z., Xie, Q., Berger, B., et~al.: High-resolution de novo structure prediction from primary sequence. BioRxiv pp. 2022--07 (2022)

\bibitem{yi2024graph}
Yi, K., Zhou, B., Shen, Y., Li{\`o}, P., Wang, Y.: Graph denoising diffusion for inverse protein folding. Advances in Neural Information Processing Systems  \textbf{36} (2024)

\bibitem{zhang2024biolip2}
Zhang, C., Zhang, X., Freddolino, P.L., Zhang, Y.: Biolip2: an updated structure database for biologically relevant ligand--protein interactions. Nucleic Acids Research  \textbf{52}(D1),  D404--D412 (2024)

\end{thebibliography}
%





\end{document}